\documentclass[10pt,twocolumn,letterpaper]{article}

\usepackage{ijcb}
\usepackage{times}
\usepackage{epsfig}
\usepackage{graphicx}
\usepackage{amsmath}
\usepackage{amssymb}

\usepackage{subfigure}
\usepackage{makecell}
\usepackage[accsupp]{axessibility} % Improves PDF readability for those with disabilities.

\ijcbfinalcopy % *** Uncomment this line for the final submission

\ifijcbfinal\fi
\begin{document}

%%%%%%%%% TITLE
\title{Direct Regression of Distortion Field from a Single Fingerprint Image}

\author{Xiongjun Guan, Yongjie Duan, Jianjiang Feng, and Jie Zhou\\
Department of Automation, BNRist, Tsinghua University, China\\
{\tt\small \{gxj21@mails.,dyj17@mails., jfeng@, jzhou@\}tsinghua.edu.cn}
}

\maketitle
\thispagestyle{empty}

%%%%%%%%% ABSTRACT
\begin{abstract}
	Skin distortion is a long standing challenge in fingerprint matching, which causes false non-matches. 
	Previous studies have shown that the recognition rate can be improved by estimating the distortion field from a distorted fingerprint and then rectifying it into a normal fingerprint.
	However, existing rectification methods are based on principal component representation of distortion fields, which is not accurate and are very sensitive to finger pose.
	In this paper, we propose a rectification method where a self-reference based network is utilized to directly estimate the dense distortion field of distorted fingerprint instead of its low dimensional representation. 
	This method can output accurate distortion fields of distorted fingerprints with various finger poses.
	Considering the limited number and variety of distorted fingerprints in the existing public dataset, we collected more distorted fingerprints with diverse finger poses and distortion patterns as a new database. Experimental results demonstrate that our proposed method achieves the state-of-the-art rectification performance in terms of distortion field estimation and rectified fingerprint matching.

\end{abstract}

%%%%%%%%% BODY TEXT

\section{Introduction}

\begin{figure}[t]
	\begin{center}
		\includegraphics[width=.95\linewidth]{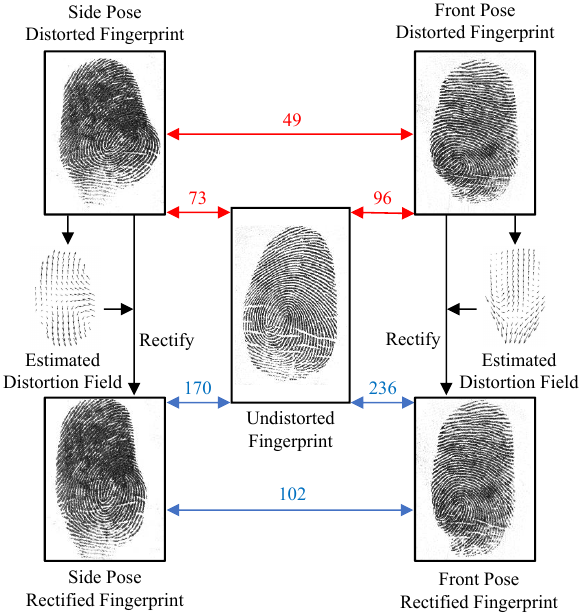}
	\end{center}
	\caption{Impact of the proposed rectification algorithm on matching performance. The red and blue numbers over arrows represent the matching scores before and after rectification calculated by VeriFinger SDK 12.0 \cite{VeriFinger}. Both distorted-undistorted and distorted-distorted matching scores are significantly improved after rectification (73 $\rightarrow$ 170, 96 $\rightarrow$ 236, 49 $\rightarrow$ 102).
	}
	\label{fig:intro}
\end{figure}

Fingerprint is one of the most important and widely used biometric traits due to its easy collection process, persistence and uniqueness \cite{maltoni2009handbook}. 
With the development of sensor technology and recognition algorithms in the past few decades, fingerprint recognition technologies developed rapidly and have been applied in various fields such as criminal investigation, mobile payment, and access control. 
Most existing fingerprint recognition algorithms extract features based on ridge and minutiae information, which are then utilized for fingerprint matching \cite{cappelli2010minutia}.
Although these algorithms achieve satisfactory performance on high-quality fingerprint images, they often fail to identify severely distorted fingerprints \cite{cappelli2005performance}. 

\begin{figure}[t]
	\begin{center}
		\begin{center}
			% \fbox{\rule{0pt}{4in} \rule{.95\linewidth}{0pt}}
			\subfigure[]{\includegraphics[width=.69\linewidth]{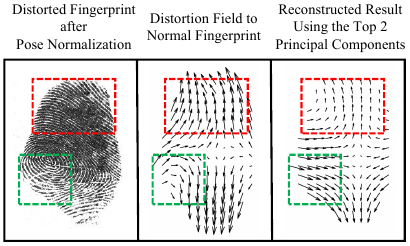}} 
			\subfigure[]{\includegraphics[width=.3\linewidth]{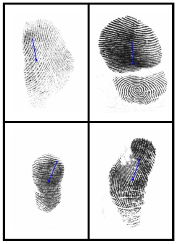}}
		\end{center}
	\end{center}
	\caption{Failure cases of PCA based rectification. (a) Distortion details are lost. Green and red rectangles represent regions where orientation or magnitude information is lost. (b) Examples where finger pose is difficult to estimate accurately. The blue arrow indicates the finger center and direction extimated by \cite{yin2020joint}. 
	}
	\label{fig:pca_ex}
\end{figure}

Since the finger pulp is curved and soft, fingerprint will deform when it is in contact with acquisition equipment, and the distortion field becomes obvious when it is subjected to lateral force or torque \cite{maceo2009qualitative}. In addition, the distortion field of the same finger is different under different pressing poses and strengths (see Figure \ref{fig:intro}), which makes the recognition of distorted fingerprints more challenging. 
In general, fingerprint distortion changes the ridge orientation, ridge frequency, and relative positions of minutiae, which increases intra-class variations among fingerprints from the same finger and thus reduces the matching performance. 
In positive identification scenario such as computer logon, distorted fingerprints will lead to false rejection, very frustrating for the user. 
In negative identification scenario, persons in the watch-list may deliberately distort their fingerprints to deceive the recognition system \cite{wein2005using,soweon2012altered},  which is a huge security risk.
Therefore, it is important but still challenging to improve the recognition performance for distorted fingerprints.

Researchers have proposed several methods to overcome the negative effects introduced by contact fingerprint distortion, which can be classified as contactless 3D imaging, distortion detection, distortion-tolerant matching, elastic registraition, and distortion rectification. Although contactless 3D fingerprints can avoid skin distortion \cite{Kumar2018,Yin2021}, this technology has not been widely used.  Distortion detection usually requires additional sensors in the fingerprint acquisition stage \cite{ratha1998effect,bolle2000system,dorai2004dynamic,fujii2010detection}, which is expensive and cannot be applied to existing fingerprint databases. Distortion-tolerant matching \cite{kovacs2000fingerprint,watson2003correlation,jea2005minutia,chen2006new,zheng2007robust,tong2008local} inevitably increases the false match rate while allowing large distortion. Elastic registration \cite{almansa2000fingerprint,bazen2003fingerprint,si2017dense,cui2021dense} needs to be performed on every matching pair which is time-consuming for identification systems. 
While given a single distorted fingerprint, distortion rectification algorithms can rectify it to remove distortion component before matching stage, as shown in Figure \ref{fig:intro}, which is more universal (no need to change the existing acquisition and matching modules) and efficient (only performed once after acquisition). Conventional rectification algorithms estimate the distortion field represented by Principal Component Analysis (PCA) \cite{si2015detection,gu2017efficient,dabouei2018fingerprint}, whose performance is highly dependent on the performance of pose normalization, and is limited by the accuracy of principle components. 
Figure \ref{fig:pca_ex} shows several failure cases where local details are lost even if pose normalization is accurate, and the error of PCA reconstruction will be larger when the finegr pose is misestimated (such as examples in (b)).
Furthermore, most of the distorted fingerprints in public Tsinghua Distorted Fingerprint (TDF) database are of front pose, high quality, and have simple distortion patterns, which are relatively easy to rectify.

In this paper, we propose a dense rectification method, which directly estimates the distortion field from a single distorted fingerprint image. 
Correlation between local and global distortion patterns is captured so as to finely estimate detailed distortion patterns, rather than a linear superposition of principle components as previous studies, which can only restore a rough distortion field.
Meanwhile, the dependence on finger pose is avoided because we directly regress the distortion field through convolution neural network, which is translation invariant.
Briefly, the main contributions can be summarized as:
\begin{itemize}
	\item We proposed an end-to-end network to directly estimate a dense distortion field without fingerprint pose alignment, instead of its low dimensional representation, from a single fingerprint;
	\item  We collected 480 videos of fingerprint distortion, including many fingerprints with diverse poses and various distortion types. Experiments were performed on this database and TDF to measure distortion estimation accuracy, matching performance, model complexity, and inference efficiency.
\end{itemize}

Experimental results on this extended dataset show that complicated distorted fingerprints can be rectified and the proposed algorithm achieves the state-of-the-art distortion rectification performance.

\section{Related Work}

Senior and Bolle \cite{senior2001improved} made the first attempt to rectify distortion for a single fingerprint. They assumed that fingerprints have a consistent ridge period, and proposed an algorithm to normalize the ridge period of fingerprint
However, the ridge period is different even in different regions of the same finger. Simply normalizing them will lose some important recognition information. 
At the same time, more distortion may be generated if only the period is adjusted regardless of orientation field.

\begin{figure*}[t]
	\begin{center}
		\begin{center}
			% \fbox{\rule{0pt}{4in} \rule{.95\linewidth}{0pt}}
			\subfigure[]{\includegraphics[width=.6\linewidth]{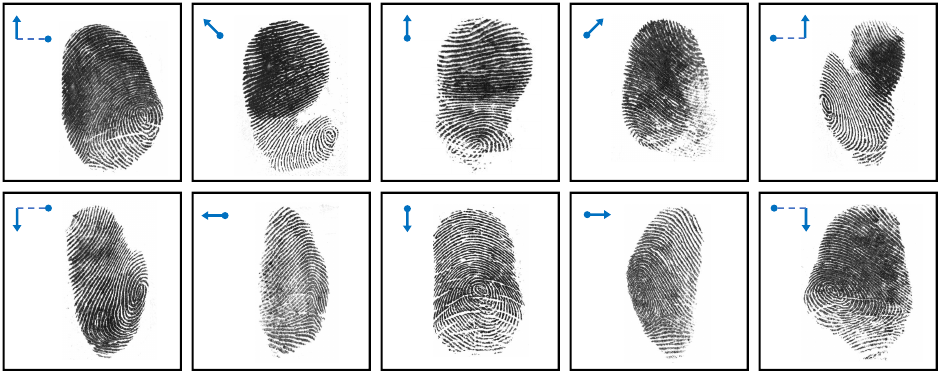}} \quad
			\subfigure[]{\includegraphics[width=.37\linewidth]{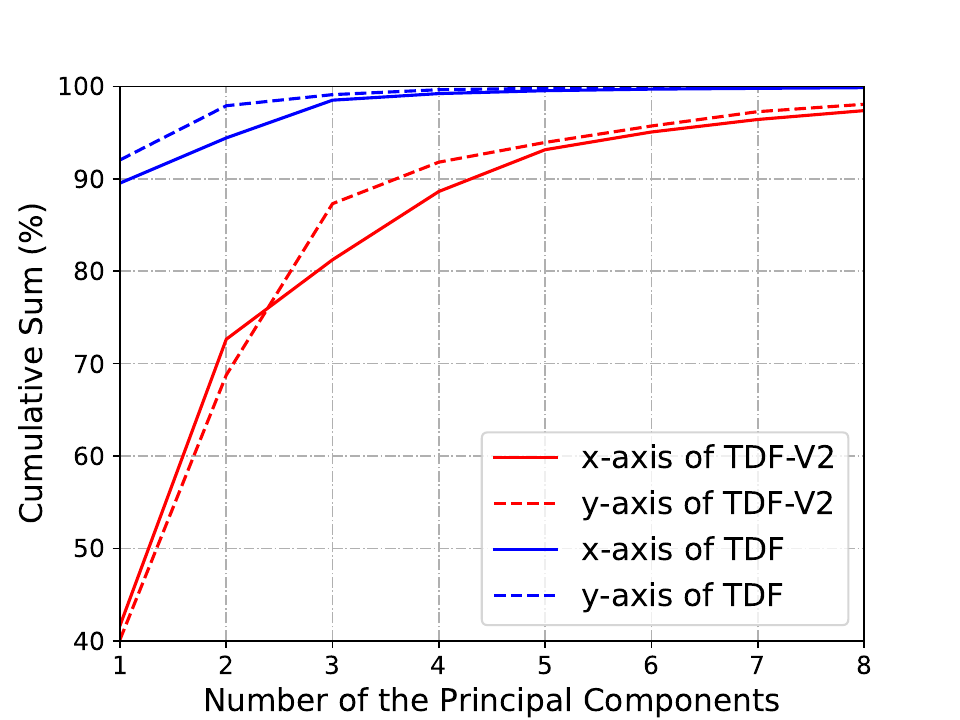}}
		\end{center}
	\end{center}
	\caption{Statistics of the distorted fingerprint database collected by us. (a) Examples of all 10 distortion types in the database. The solid blue circle represents finger of front pose, the arrow represents the direction of force when pressing, and the dotted line represents finger of side pose. (b) Cumulative sum of the main distortion patterns in TDF-V2. The numbers on abscissa represent descending ranks of the top eight principal components in two axes. Finger pose of each fingerprint is normalized in advance.}
	\label{fig:data}
\end{figure*}

Si \etal \cite{si2015detection} conducted a further study of distortion rectification.
They collected the TDF database, which contains 320 videos of distorted fingerprints, and extracted the main distortion patterns by PCA.
They constructed a dictionary of generated distorted fingerprints in the off-line stage and estimated the distortion field of an input fingerprint by nearest neighbor search based on ridge orientation and periods.
Gu \etal \cite{gu2017efficient} improved the nearest neighbor search step in \cite{si2015detection} by utilizing support vector regression to predict the distortion parameters, which greatly improved the efficiency.
Dabouei \etal \cite{dabouei2018fingerprint} used a deep convolution neural network to directly estimate the coefficients of principal components from an input fingerprint. Experiments show that these methods can improve the matching accuracy of most distorted fingerprints, but there are still some limitations: (1) normalizing the location and pose of fingerprint in advance is required, which making these algorithms fail to deal with fingerprints whose poses cannot be accurately estimated, such as small areas, low quality, or large roll or pitch angles; (2) only principal distortion is estimated, thus distorted fingerprints with complex distortion patterns cannot be accurately rectified.

\begin{figure*}[t]
	\begin{center}
		\includegraphics[width=0.9\linewidth]{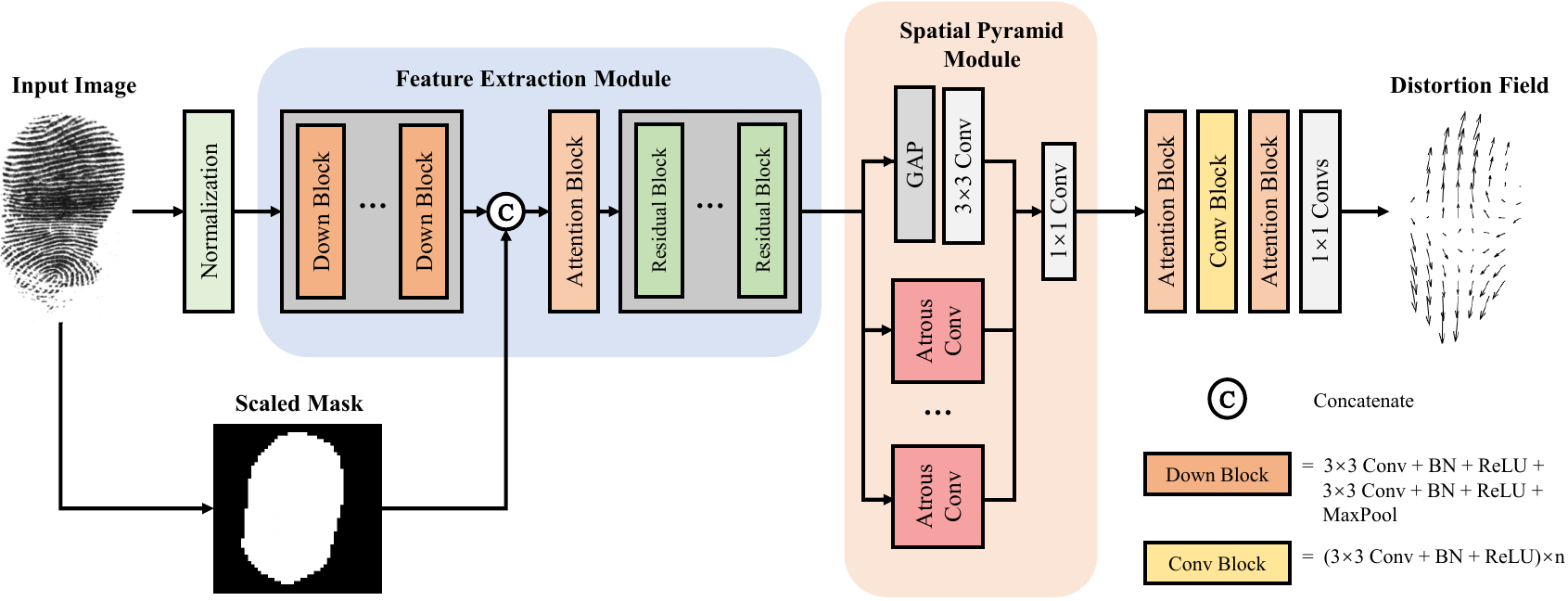}
	\end{center}
	\caption{Structure of the proposed distortion estimation network. The network includes a downsampling module and a residual module for feature extraction, a spatial pyramid module for fusing multi-scale feature information, and a convolution regression module for predicting dense distortion field. For an input distorted fingerprint, the network takes the $16 \times 16$ pixel block as a unit and gives its corresponding distortion field (displacement from the input image to the rectification target).}
	\label{fig:network}
\end{figure*}

\section{Method}
In this paper, we aim to predict a dense distortion field ditectly from a single fingerprint by utilizing multi-scale feature information. 
The proposed network structure is shown in Figure \ref{fig:network}, which takes the distorted fingerprint and its scaled mask as input, and outputs the two-dimensional distortion field to the rectification target. The rectified fingerprint is obtained by shifting the input image pixel by pixel according to the prediction result of the network.
In section \ref{sec:data}, the database is introduced, Section \ref{sec:gt} discribes how to calculate the ground truth of distortion field and perform the data augmentation, and Section \ref{sec:network} presents the network architecture and loss functions.

\subsection{Distorted fingerprints database} \label{sec:data}
To take full advantage of the learning capability of neural networks, a large amount of distorted fingerprint data needs to be provided. Although distorted fingerprints can be imitated by synthesizing distortion fields and applying them on normal fingerprints, the simulated result is not guaranteed to be realistic.
Therefore, in addition to using the existing TDF database, we also collected more abundant distorted fingerprints.
The TDF established by Si \etal \cite{si2015detection} has 320 distorted fingerprints. In the collection process of this database, fingers were first pressed in the front pose, and then deformed by horizontal or vertical force or torque. Considering that the common fingerprint distortion in practice is usually caused by unidirectional force and may occur with various poses, we collected additional 480 distorted fingerprint videos generated by one-way rubbing in different directions on the front or side finger pose, instead of torque and two-stage force types in TDF. A Frustrated Total Internal Reflection (FRIR) fingerprint scanner was used to obtain fingerprint sequences. Totally 10 distortion types of videos for each finger are sampled at 30Hz with a resolution of 500 ppi, and the duration of each video is about 3s. 
Examples of 10 distortion fingerprint types and the principal component distribution of all data are shown in Figure \ref{fig:data}. It can be seen that with different initial poses and pressing directions, different distortion patterns will be produced (it takes about 6-8 principal components to represent the distortion field well, instead of 2 for TDF).
We merged the above two distorted fingerprint databases, and named the new database as TDF-V2 for convenience, and use 640 videos for training and the remaining 160 for testing.  
The images are cropped to $512\times512$, and the finger pose of each fingerprint is normalized using the method in \cite{yin2020joint}.

\subsection{Training data preparation} \label{sec:gt}
We get distorted fingerprints from TDF-V2 as input and the distortion field from the distorted fingerprint to the normal fingerprint as the ground truth.
Similar to \cite{si2015detection,gu2017efficient,dabouei2018fingerprint}, we take the initial frame as the normal fingerprint and the end frame as the distorted fingerprint, and obtain the displacement between them by pairing minutiae points in adjacent frames and performing thin-plate spline interpolation. 
Instead of simply computing difference as existing methods, given a distorted fingerprint ${\rm D}$ and its corresponding normal fingerprint ${\rm N}$, we first align the fingerprint pairs rigidly and then extract non DC (Direct Current) components in distortion as follows:
\begin{equation}
	\begin{aligned}
		R^{{\rm ND}}, t^{{\rm ND}}&=\underset{R, t}{\arg \min }\left[\left(R \cdot P^{{\rm N}}+t\right)-P^{{\rm D}}\right] \cdot M^{\rm D} ,\\
		F_{i}^{{\rm DN}}&=\left(R^{{\rm ND}} \cdot P_{i}^{{\rm N}}+t^{{\rm ND}}\right)-P_{i}^{{\rm D}} ,
	\end{aligned}
	\label{eq:constraint}
\end{equation}
where $F_i^{{\rm DN}}$ is the displacement between the $i$th paired points, $P$ is the set of coordinates, $M$ is the mask, $R$ and $t$ are rigid rotation and translation matrices.
With the strong constraint proposed above, the situation of multiple solutions due to rigid transformations is avoided, and unnecessary DC component caused by finger translation or rotation during acquisition will not be introduced.

In order to increase the data size, the training set is augmented by mirror reflection and rotation. 
Each distorted fingerprint is flipped horizontally and rotated by 90, 180 and 270 degrees.
Through these augmentation methods, the data is $8$ times bigger than the original.

\subsection{Network architecture}\label{sec:network}
Previous rectification methods \cite{si2015detection,gu2017efficient,dabouei2018fingerprint} have proved that extracting and learning distortion patterns from fingerprint video is feasible and effective.
In addition, the dense fingerprint registration method proposed by Cui \etal \cite{cui2021dense} also shows good results in reducing the distortion between paired fingerprints, which predicts the dense distortion field by a siamese block and an encoder-decoder.
Inspired by these methods, although there is no paired input fingerprint in single fingerprint rectification, we still construct the reference relationship between the features on different spatial scales, so as to learn the patterns of dense distortion field.

Figure \ref{fig:network} illustrates the structure of our dense distortion estimation network. 
Apart from the distorted fingerprint, the network also takes fingerprint mask as input due to the mask-related constraints used in Equation \ref{eq:constraint} when generating the ground truth. Mask here is the largest connected domain of the segmentation result, which is obtained by calculating the gradient and setting a  threshold for the input image. 
In order to minimize the interference of different image grayscales, fingerprints are normalized before entering the network. 

The network can be divided into three parts. In the feature extraction module, high-level features are obtained through 4 downsampling blocks with a convolution kernel of 3, and then concated with the scaled mask. 
It should be noted that we add the mask after downsampling blocks instead of before, because it is used to constrain the estimation of the distortion and not helpful for analyzing the regional features.
To better focus on the information of multiple channels, a coordinate-sensitive channel attention block \cite{hou2021coordinate} is adopted, which decomposes the channel attention into two parallel coordinate-dependent 1D feature encodings to help the model locate and identify interest regions. After assigning weights to channels, several residual blocks are concatenated to better extract and integrate feature information.
Since fingers have a continuous irregular elastic surface, the distortion field of the distorted fingerprint is usually relatively uniform in a large area, and it is locally different at the same time.
In order to capture those contextual information at multiple scales, global average pooling and atrous convolution blocks with different dilation rates are parallelized in the spatial pyramid module, inspired by \cite{chen2018encoder}.
In the last part, attention blocks and convolution blocks are used to regress the distortion field.

The output of the network regards the $16\times16$ pixel block as a displacement unit, which is fine enough to correct the distortion of the fingerprint (similarly, it is usually scaled to $1/8$ size of the original image in the orientation estimation task) and make the network lightweight at the same time. The estimated distortion field is interpolated to the original size and then applied to the rectification step.

The training loss consists of two parts: the regression loss $\mathcal{L}_{{\rm reg}}$ between the estimated distortion field $F^{{\rm est}}$ and the ground truth $F^{{\rm gt}}$, and the smoothing loss $\mathcal{L}_{{\rm smo}}$ according to the gradient of $F^{{\rm est}}$
\begin{align}
	\mathcal{L}_{{\rm reg}}&=\frac{1}{\sum_{i,j}M^{\rm D}} \left\|\left( F^{{\rm est}}-F^{{\rm gt}}\right)\cdot M^{\rm D}\right\|_{2}^{2}, \label{eq:reg}\\
	\mathcal{L}_{{\rm smo}}&= \frac{1}{h\cdot w} \sum_{i,j}\left(\sum_{c=1}^{2} \left( \left|\nabla F_{x}^{ {{\rm est }}}\right|^{2}+\left|\nabla F_{y}^{ {{\rm est} }}\right|^{2}\right)\right),\\
	\mathcal{L}&=\mathcal{L}_{{\rm reg}}+\lambda_{{\rm smo}}\cdot \mathcal{L}_{{\rm smo}},
\end{align}
where $i$ and $j$ are the coordinates corresponding to the height $h$ and weight $w$ of the image, $M^{\rm D}$ is the mask of distorted fingerprint, the trade-off weight $\lambda_{smo}$ is set as $1.0$ to balance these loss functions.

\section{Experiments}
As mentioned in Section \ref{sec:data}, we use 160 distorted fingerprint videos in TDF-V2 for testing, called TDF-V2\_T. 
The last and first frame of each video were selected as the distorted fingerprint and its corresponding normal fingerprint respectively. 
For fingers with multiple videos, only the initial frames with large differences are retained to avoid the interference of duplicate data. In this way, a total of 710 genuine matches are obtained.
We compare the proposed method with state-of-the-art rectification algorithms in terms of distortion estimation accuracy, rectified fingerprint matching performance, model complexity, and inference efficiency.
Since the distortion in this dataset is more complex than TDF (as shown in Figure \ref{fig:data} (b)), PCA based methods \cite{gu2017efficient,dabouei2018fingerprint} use the top 8 principal components instead of 2 in our reimplementation. 
Considering that the encoder-decoder structure performs well in pair-wise distortion field estimation \cite{cui2021dense}, we also add U-Net \cite{ronneberger2015u} as a baseline for comparison.
The method proposed by Si \etal \cite{si2015detection} is not included since the algorithm is highly dependent on the dictionary size and the retrieval process is significantly time-consuming, making it difficult to apply in practice, and previous studies showed that its performance has been exceeded by \cite{gu2017efficient,dabouei2018fingerprint}.

\begin{figure}[h]
	\begin{center}
		\includegraphics[width=.95\linewidth]{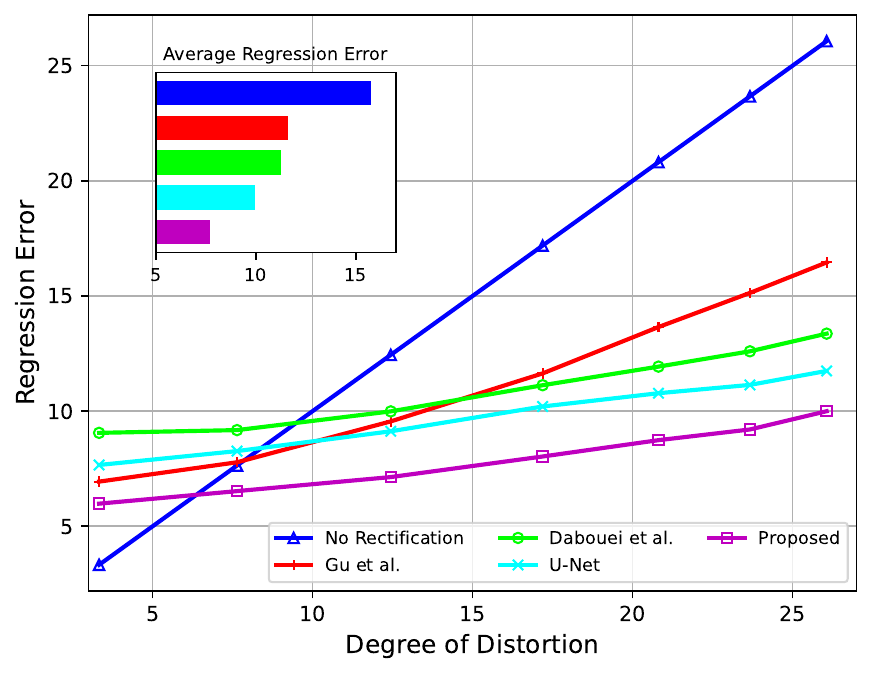}
	\end{center}
	\caption{Regression error respectively of different distortion field estimation algorithms on TDF-V2\_T. Color legend of bar chart and line chart is consistent. The degree of distortion is divided into seven intervals.}
	\label{fig:dis}
\end{figure}

\begin{table}[t]
	{\small
	\caption{Ablation experiments of the proposed network on TDF-V2\_T. `Mask-A': mask is concatenated after normalization. `Mask-B': mask is concatenated as Figure \ref{fig:network}.}
	\label{tab:ablation}
	\begin{center}
		\begin{tabular}{|c c c c|c|}
			\hline
			\makecell[c]{Spatial\\Pyramid\\module} & \makecell[c]{Attention\\Block} & \makecell[c]{Mask-A} & \makecell[c]{Mask-B}  & $\mathcal{L}_{{\rm reg}}^{\rm R}$	\\
			\hline\hline
			-			&-			&-			&-			& 10.20\\ 
			$\surd$ 	&- 			&- 			&- 			& 8.27\\
			$\surd$ 	&$\surd$	&- 			&- 			& 8.03\\
			$\surd$ 	&- 			&$\surd$	&- 			& 8.13\\
			$\surd$ 	&- 			&- 			&$\surd$	& 8.13\\
			$\surd$ 	&$\surd$	&$\surd$	&- 			& 7.81\\ \hline
			$\surd$ 	&$\surd$	&- 			&$\surd$	& {\bf7.69}\\
			\hline
		\end{tabular}
	\end{center}}
\end{table}

\setcounter{figure}{6}
\begin{figure*}[h]
	\begin{center}
		\includegraphics[width=.98\linewidth]{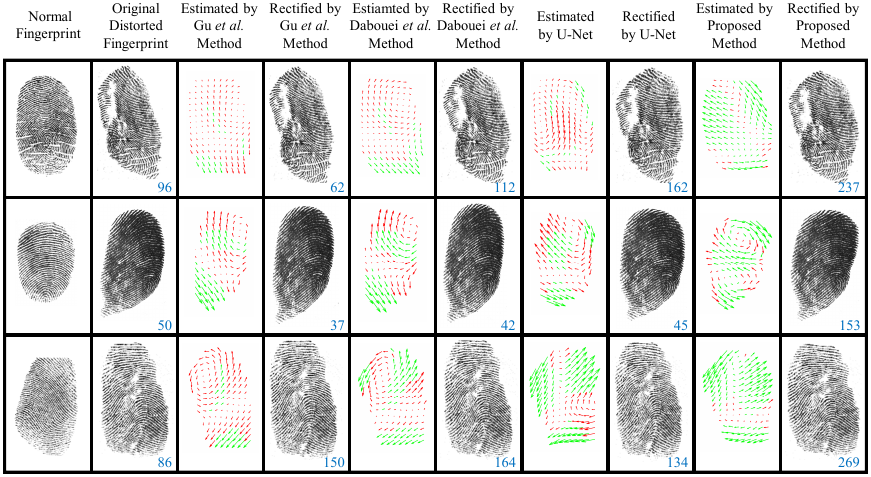}
	\end{center}
	\caption{Examples of different rectification methods. The blue numbers are matching scores between the normal fingerprint and the distorted one (original and rectified by different methods) calculated by VeriFinger.  The green vectors in the estimated distortion field indicate the correctly estimated distortion elements, while the red vectors for wrong.
	}
	\label{fig:example}
\end{figure*}

\subsection{Distortion estimation accuracy}

The distortion estimation accuracy is evaluated through the regression error  $\mathcal{L}_{{\rm reg}}^{\rm R}$ (in pixel) between the prediction result of proposed network and the ground truth, which is similar as Equation (\ref{eq:reg}) except that the square $\left\|\cdot\right\|_2^2$ is replaced by the square root $\left\|\cdot\right\|_2$. 
We regard amplitude of each point in the distortion field (without rectification) as its degree of distortion, and divide the complete mask into different sub regions pixel by pixel according to it. Regression error of each region is calculated separately, so as to specifically observe the estimation accuracy under different degrees of distortion.
As shown in Figure \ref{fig:dis}, dense estimation methods (U-Net and ours) perform better than PCA based methods \cite{gu2017efficient,dabouei2018fingerprint} on large distoriton regions, which proves that directly regressing the distortion field is more refined. Proposed method significantly outperforms other methods in both regions with large and small distortions. Even for regions with slight distortion, our network still performs appropriate rectification. More examples will be provided in Section \ref{sec:performance} to illustrate this point.
Moreover, ablation experiments are given in Table \ref{tab:ablation} to verify the value of specific modules. For fairness, we fine-tune the network parameters to ensure that the model sizes are roughly the same.

\subsection{Matching performance} \label{sec:performance}
To quantitatively evaluate the performance of these distortion rectification methods in a complete fingerprint recognition system, we further evaluated the matching performance using fingerprints after distortion rectification. Matching scores are computed by a commercial software VeriFinger SDK 12.0 \cite{VeriFinger}. 

\setcounter{figure}{5}
\begin{figure}[h]
	\begin{center}
		\includegraphics[width=.95\linewidth]{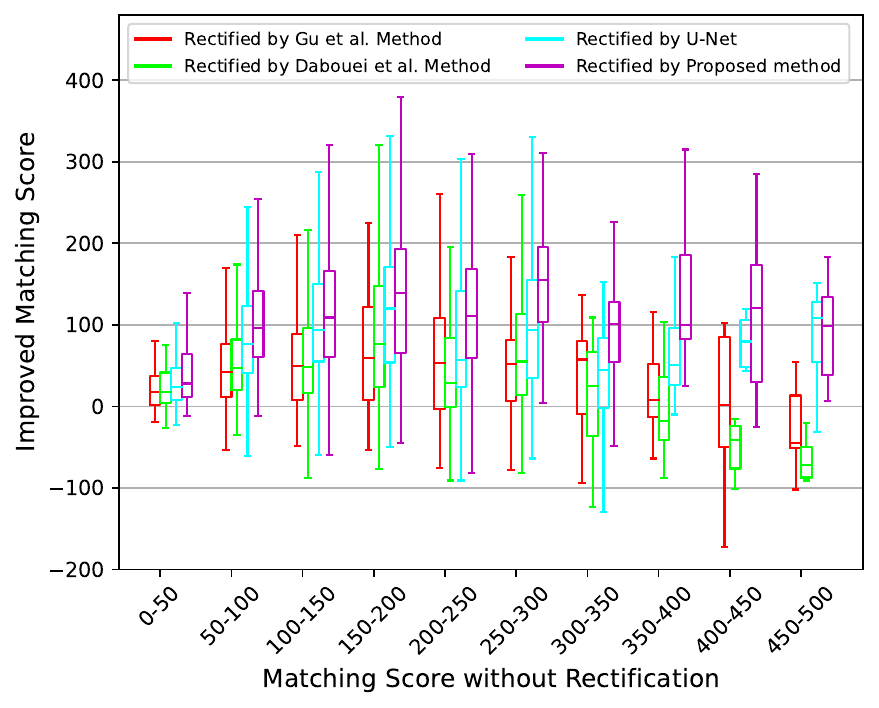}
	\end{center}
	\caption{Boxplot of VeriFinger matching score improvements after rectification  on TDF-V2\_T. The horizontal line inside the box is the median of the corresponding data.
	}
	\label{fig:improvement}
\end{figure}
\setcounter{figure}{7}

\begin{figure*}[h]
	\begin{center}
		\begin{center}
			% \fbox{\rule{0pt}{4in} \rule{.95\linewidth}{0pt}}
			\subfigure[]{\includegraphics[width=.45\linewidth]{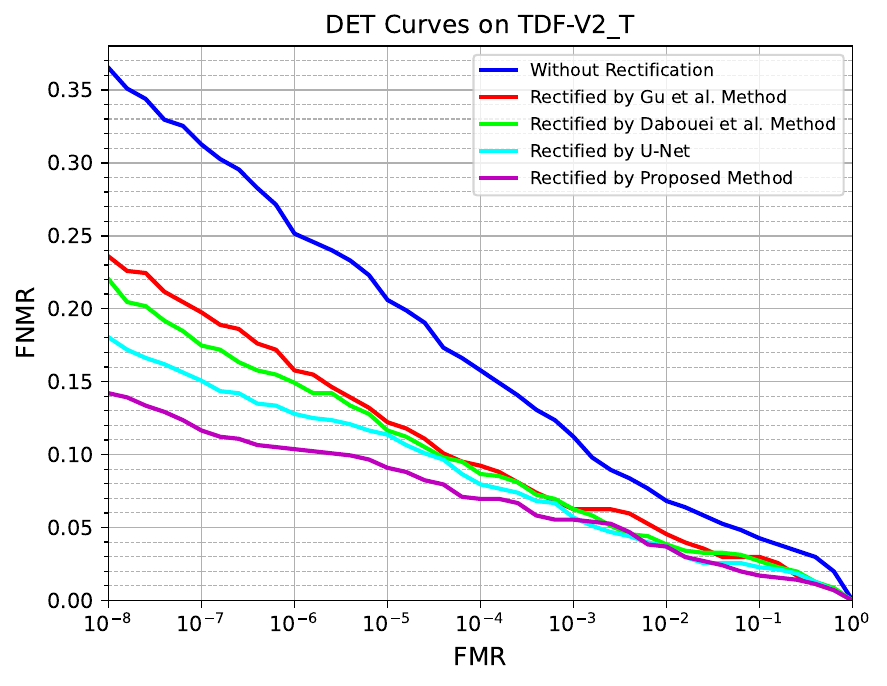}} \quad
			\subfigure[]{\includegraphics[width=.45\linewidth]{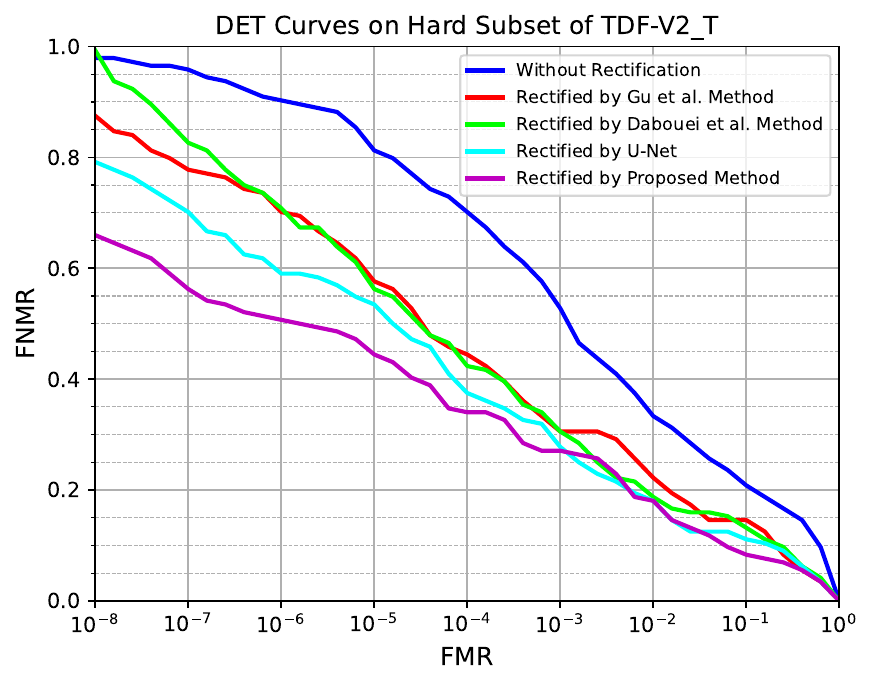}}
		\end{center}
	\end{center}
	\caption{DET curves by VeriFinger matcher with rectification algorithms on (a) TDF-V2\_T, and (b) hard subset of TDF-V2\_T.}
	\label{fig:det}
\end{figure*}

We first compare the matching score improvement of different rectification methods on TDF-V2\_T. Genuine match scores are calculated for each distorted fingerprint and its corresponding normal fingerprints.
Only the pairs with original matching score (without rectification) lower than 500 are shown in Figure \ref{fig:improvement}, since matching scores over 500 are sufficiently high for genuine match. From the distribution we can see that the proposed method leads prominently in each segment. PCA based methods do not work well in cases where the original matching scores are high (meaning that the fingerprint distortion is slight), because it is a global distortion and may incorrectly rectify some local areas.

\begin{table}[h]
	\caption{Model size and efficiency of different rectification methods for processing a $512\times512$ distorted fingerprint in TDF-V2\_T.}
	\label{tab:cost}
	\begin{center}
		\begin{tabular}{|l|c|c|}
			\hline
			Methods                      		& Param(MB) & Time(s) \\
			\hline\hline
			Gu \etal \cite{gu2017efficient}		& 107.1	& 5.13	\\
			Dabouei \etal \cite{dabouei2018fingerprint}	& 240.9	& 0.39	\\
			U-Net \cite{ronneberger2015u}		& 51.8	& {\bf0.33}	\\
			\hline
			Proposed                               	& {\bf 45.0} &  0.37 \\
			\hline
		\end{tabular}
	\end{center}
\end{table}

Three examples from TDF-V2\_T are given in Figure \ref{fig:example} to compare the rectified results. For a displacement vector $v$ in the distortion field whose angle is $\theta$, we regard it as a wrong estimate when it satisfies $\lvert \theta^{\rm est}-\theta^{\rm gt} \rvert > 45 ^{\circ}$ or $\left\| v^{\rm est} - v^{\rm gt} \right\|_2 / \min\{\left\|v^{\rm est}\right\|_{2},\left\|v^{\rm gt}\right\|_{2}\} > 1.2$ and mark it in red (others in green).
Existing methods expose the following problems:  (1) hard
to deal with fingerprints whose pose cannot be accurrately normalized (large roll angle and incomplete central area in line 1); (2) only significant distortion in one direction is concerned (transverse distortion is ignored in line 2); (3) cannot understand complex distortion (line 3), while our method works well in these cases.

To further evaluate the rectification performance in recognition task, Detection Error Tradeoff (DET) curves are depicted in Figure \ref{fig:det}.
Same as previous experimental settings \cite{si2015detection,gu2017efficient,dabouei2018fingerprint}, we only use genuine match scores because the match score of VeriFinger has been designed to map the false match rate (FMR). This allows us to measure the false non-match rates (FNMR) at lower FMRs despite the limited number of imposter matches.
Along with the TDF-V2\_T dataset, rectification methods are also tested on its hard subset (144 genuine matches in total, which are constructed by selecting genuine pairs with low matching scores).  It can be observed from these curves that our method exceeds other rectification methods on matching performance.

\subsection{Model complexity and efficiency}

Table \ref{tab:cost} shows the model size and efficiency of different rectification algorithms. Parameters of \cite{gu2017efficient} and \cite{dabouei2018fingerprint} contain the principal distortion patterns extracted by PCA. Time in this table is the average time from inputting a $512\times512$ distorted fingerprint in TDF-V2\_T to outputting its rectified result. It can be seen that our method has the smallest number of parameters and there is little difference in time cost except for Gu \etal method, in which additional extraction of orientation and period is required.
 All rectification algorithms are implemented in Python on a computer with a NVIDIA GeForce RTX 2080 Ti GPU and a 1.2 GHz CPU.

\section{Conclusion}
In this paper, we propose a fingerprint distortion rectification algorithm where the distortion field of a single fingerprint is directly regressed. 
Self-reference relationship is constructed in the proposed network to finely estimate the detailed distortion patterns, instead of a sparse combination of principle components.
Moreover, the performance of our proposed method does not depend on pose normalization, which is unreliable for fingerprints with several distortion and non-frontal poses.
More distorted fingerprints with diverse poses and various distortion types are collected to make a more challenging and diverse distorted fingerprint database.
Experiments show our proposed method outperforms state-of-the-art rectification algorithms.

The limitation of the method is that the fingerprints we use for training are relatively clean, thus it may not be very effective in the case of low image quality (such as complex background or very incomplete fingerprints). Additionally, PCA based methods perform better when dealing with some typical distortions because of stronger global priors. In the future we will focus on combining prior information with self-reference to estimate distortion field, and extend our method to latent fingerprints which are affected by both distortion and strong noise.

\section{Acknowledgment}
This work was supported in part by the National Natural Science Foundation of China under Grant 61976121.

{\small
\bibliographystyle{ieee}
\bibliography{egbib}
}

\end{document}